\documentclass{article}

\usepackage{microtype}
\usepackage{graphicx}
\usepackage{subcaption}
\usepackage{booktabs} 
\usepackage{algorithm}
\usepackage{algorithmic}
\usepackage{mwe} 
\usepackage{xcolor}
\usepackage{pifont}

\newcommand{\safeincludegraphics}[2][]{%
    \IfFileExists{#2}{\includegraphics[#1]{#2}}{\fbox{Missing figure: #2}}%
}

\usepackage{hyperref}

\newcommand{\citelink}[2]{\hyperlink{cite.#1}{#2}\nocite{#1}}

\usepackage[preprint]{icml2026}

\usepackage{amsmath}
\usepackage{amssymb}
\usepackage{mathtools}
\usepackage{amsthm}

\usepackage[capitalize,noabbrev]{cleveref}

\theoremstyle{plain}

\theoremstyle{definition}

\theoremstyle{remark}

\usepackage[textsize=tiny]{todonotes}

\newcommand{\cmark}{\textcolor{blue}{\ding{51}}} 
\newcommand{\xmark}{\textcolor{red}{\ding{55}}}   
\icmltitlerunning{SiDGen: Structure-informed Diffusion for Generative Modeling of Ligands for Proteins}

\begin{document}

\twocolumn[
  \icmltitle{SiDGen: Structure-informed Diffusion for Generative Modeling of Ligands for Proteins}



  \icmlsetsymbol{equal}{*}

  \begin{icmlauthorlist}
    \icmlauthor{Samyak Sanghvi}{yyy}
    \icmlauthor{Nishant Ranjan}{yyy}
    \icmlauthor{Tarak Karmakar}{sch}
  \end{icmlauthorlist}

  \icmlaffiliation{yyy}{Department of Computer Science, Indian Institute of Technology, Delhi, India}
  \icmlaffiliation{sch}{Department of Chemistry, Indian Institute of Technology, Delhi, India}

  \icmlcorrespondingauthor{Samyak Sanghvi}{samyakssanghvi@gmail.com}
  \icmlcorrespondingauthor{Tarak Karmakar}{tkarmakar@chemistry.iitd.ac.in}


  \vskip 0.3in
]



\printAffiliationsAndNotice{}  

\begin{abstract}
Structure-based drug design (SBDD) faces a fundamental \textbf{scaling fidelity dilemma}: rich pocket-aware conditioning captures interaction geometry but can be costly, often scales quadratically ($O(L^2)$) or worse with protein length ($L$), while efficient sequence-only conditioning can miss key interaction structure. We propose \textbf{SiDGen}, a structure-informed discrete diffusion framework that resolves this trade-off through a \textbf{Topological Information Bottleneck (TIB)}. SiDGen leverages a learned, soft assignment mechanism to compress residue-level protein representations into a compact bottleneck enabling downstream pairwise computations on the coarse grid ($O(L^2/s^2)$). This design reduces memory and computational cost without compromising generative accuracy. Our approach achieves state-of-the-art performance on CrossDocked2020 and DUD-E benchmarks while significantly reducing pairwise-tensor memory. SiDGen bridges the gap between sequence-based efficiency and pocket-aware conditioning, offering a scalable path for high-throughput structure-based discovery.
\footnote{The code with instructions can be found at \href{https://github.com/SamyakSS83/SiDGEN}{https://github.com/SamyakSS83/SiDGEN/}}
\end{abstract}

\section{Introduction}

The computational search for novel therapeutics involves navigating a chemical space of approximately $10^{60}$ drug-like molecules. Within this space, structure-based drug design (SBDD) aims to generate ligands that exhibit high binding affinity and specificity for a target protein pocket. Recent advancements in deep generative modeling, particularly denoising diffusion models \citep{hoogeboom2022equivariant, vignac2022digress}, have shown significant promise in generating chemically diverse and geometrically accurate molecules. However, when applied to protein-conditioned generation, these models face a critical \textbf{scaling-fidelity dilemma}.

\paragraph{The Scaling-Fidelity Dilemma.} High-fidelity structural generative models typically operate on atomic point clouds or fine-grained graph representations of protein pockets. \citep{guan2023targetdiff, peng2022pocket} To capture the intricate complementarity between a ligand and its receptor, these models employ E(3)-equivariant neural networks or dense transformer architectures that compute all-to-all interactions. Although expressive, the memory and computational costs of such operations scale quadratically ($O(L^2)$) or even cubically with the number of protein residues $L$. This `computational wall' limits the ability of practitioners to process large proteins or handle multi-target discovery campaigns where throughput is essential.

In contrast, sequence-centric models utilize pre-trained protein language models (pLMs) such as ESM-2 \citep{lin2023evolutionary} to provide efficient, linear-time conditioning ($O(L)$). Although these pLMs capture rich evolutionary and implicit structural semantics, they are often used as `black-box' sequence embeddings that fail to explicitly model the 3D topological information of a specific binding pocket. As a result, such approaches often generate molecules that are chemically valid yet `target-agnostic', lacking the geometric specificity necessary for high affinity and selective binding.

\paragraph{SiDGen: Resolving the Dilemma \textit{via} Topological Information Bottlenecks.} In this work, we propose \textbf{SiDGen} (Structure-informed Diffusion Generator), a framework that bridges the gap between sequence efficiency and structural awareness. Our approach is based on the observation that while individual residue interactions are fine-grained, the \emph{complementarity} of a binding pocket is often governed by persistent topological clusters, arrangements of secondary structure elements, and conserved motifs that define the geometry of the binding pocket.\cite{guharoy2010conserved}

We formalize this observation through the concept of a \textbf{Topological Information Bottleneck (TIB)}. SiDGen learns to pool dense residue-level features into a compact, coarse-grained bottleneck representation $z \in \mathbb{R}^{K \times d}$, where $K \ll L$. Using a DiffPool-inspired soft assignment mechanism \citep{ying2018hierarchical}, the model dynamically identifies a subset of residues that form critical interaction units. This aggregation reduces the complexity of subsequent generative operations from $O(L^2)$ to $O(K^2)$, enabling scalable processing of large proteins (\textit{e.g.}, $L=2000$) on consumer-grade hardware while preserving the essential structural context.

\paragraph{Chemformer-style Discrete Diffusion.} To ensure that the generated molecules are both chemically robust and target-responsive, SiDGen operates in the discrete space of SMILES strings using a Chemformer-style tokenizer and transformer denoiser architecture \citep{irwin2022chemformer}. Masked diffusion in the SMILES space allows the model to leverage strong chemical priors while being precisely conditioned on the topological bottleneck. We augment this process with training-time stability components, including an in-loop validity penalty and a curriculum-based noise schedule.

\paragraph{Summary of Contributions.}
\begin{itemize}
    \item We introduce the \textbf{Topological Information Bottleneck} as a principled framework for efficient structural conditioning in SBDD.
    \item We propose \textbf{SiDGen}, integrating a hierarchical soft-assignment structural encoder (DiffPool-inspired) with a Chemformer-style discrete diffusion backbone, achieving $O(K^2)$ scaling.
    \item We provide an exhaustive empirical evaluation on \textbf{MOSES}, \textbf{CrossDocked2020}, and \textbf{DUD-E} benchmarks, including a comparison with recent state-of-the-art models such as \textbf{RxnFlow} \citep{powers2025rxnflow}, \textbf{CIDD} \citep{chen2025cidd}, and \textbf{MolJO} \cite{qiu2025piloting}.
    \item We demonstrate a superior \textbf{compute-quality Pareto front}, showing that SiDGen maintains high enrichment factors while using $\sim$94\% less end-to-end VRAM than full structural conditioning baselines.
\end{itemize}

The resulting SiDGen framework achieves a Vina Docking score of -9.81 kcal/mol and an enrichment factor EF@1\% of 10.26 on the DUD-E benchmark. These results demonstrate that high-fidelity pocket-aware ligand generation can be achieved without the quadratic scaling overhead associated with fine-grained structural modeling.

\begin{figure*}[t]
    \centering
    \fbox{
        \includegraphics[width=0.95\linewidth]{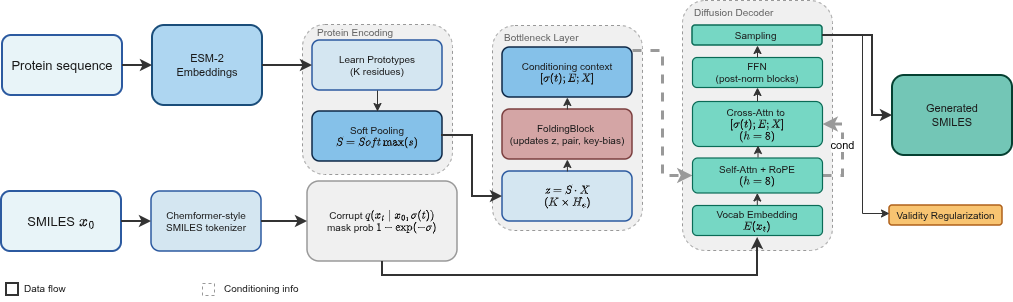}
    }
    \caption{Overview of the SiDGen architecture. A target protein is processed through a hierarchical structural encoder to produce a Topological Information Bottleneck $z$, which conditions a Chemformer-style discrete diffusion process in SMILES space.}
    \label{fig:architecture}
\end{figure*}

\section{Background}
\label{sec:background}

\subsection{Discrete Diffusion in Chemical Space}
\label{sec:bg_diffusion}
Diffusion models for discrete data, such as SMILES strings or molecular graphs, differ from their continuous counterparts in their corruption process. Given a ligand $x_0 \in \mathcal{V}^M$ from a vocabulary $\mathcal{V}$ of size $V$, the forward process $q(x_t \mid x_{t-1})$ is defined by a transition matrix $\mathbf{Q}_t \in [0,1]^{V \times V}$. The probability of being in state $j$ at time $t$ given state $i$ at time $0$ is:
\begin{align}
    q(x_t = j \mid x_0 = i) = [x_0 \bar{\mathbf{Q}}_t]_j
\end{align}
where $\bar{\mathbf{Q}}_t = \prod_{s=1}^t \mathbf{Q}_s$. 
In practice, $\mathbf{Q}_t$ is often parametrized as a mixture of identity and uniform noise distribution, resulting in a closed form expression for $q(x_t\mid x_0)$.  

In \textbf{SiDGen}, we employ a masking transition in which a token is replaced by a special [MASK] token with probability $\sigma(t)$. This transition preserves the structural identity of unmasked tokens, allowing the model to focus on denoising specific chemical motifs within the protein's structural context.



\section{Related Work}
\label{sec:related_work}
\subsection{Generative Models for SBDD}
The field of structure-based \textit{de novo} design has transitioned from search-based methods to deep generative modeling. The early benchmarks were dominated by autoregressive models and VAEs \citep{gomez2018automatic}. More recently, diffusion models have set new benchmarks for 3D pose generation. \textbf{TargetDiff} \citep{guan2023targetdiff} and \textbf{Pocket2Mol} \citep{peng2022pocket} utilize equivariant GNNs to model protein-ligand interactions in continuous space. Although effective, these models require full structural awareness, leading to quadratic scaling issues.

Recently, a new wave of models has emerged to improve geometric feasibility and practical drug-likeness. \citet{wang2024molcraft} propose \textbf{MolCRAFT} -- an SBDD generative model that operates in a continuous parameter space with a noise-reduced sampling strategy to improve pose stability and binding affinity. \textbf{MolJO} \citep{qiu2025piloting} studies the coupling between discrete (2D topology) and continuous (3D geometry) modalities and proposes an improved noise scheduling strategy to improve interaction modeling and pose quality. In parallel, \textbf{RxnFlow} \citep{powers2025rxnflow} incorporates reaction-aware constraints, while \textbf{CIDD} \citep{chen2025cidd} leverages collaboration between physics-based structure signals and large language models. SiDGen is complementary: rather than redesigning the 3D generator, we target the \emph{scaling bottleneck} by compressing the protein conditioning signal into a learnable topological bottleneck.

\subsection{Discrete Denoising and Language Models}
The success of transformers in chemical space \citep{irwin2022chemformer, ross2022molformer} has led to several SMILES-based diffusion models. \textbf{LDMol} \citep{chang2024ldmol} demonstrates that operating in string space allows models to capture long-range chemical dependencies that are often missed by graph-based methods. However, conditioning these models on 3D structures without linear-time approximations remains a challenge. Our work integrates the learned pooling of the protein representation with an efficient SMILES-diffusion backbone.

\subsection{Efficiencies in Protein Modeling}
Efforts to scale protein modeling include MSA-subsampling and token-dropping in ESM-like models. However, these are often fixed heuristics. Learnable pooling approaches have been successful in fold classification, but have seen limited use as conditioning bottlenecks for generative tasks. Our \textbf{Topological Information Bottleneck (TIB)} is inspired by these hierarchical representations but optimized specifically for ligand complementarity.

\section{Methodology}

\subsection{Hierarchical Structural Diffusion (HSD)}
\label{sec:meth_hsd}
We formulate protein-conditioned ligand generation as a conditional diffusion process $p(x\mid\mathcal{P})$, where $x$ is a ligand SMILES string, and $\mathcal{P} \in \mathbb{R}^{K \times d_p}$ is the target protein representation (ESM-2 residue embeddings). To resolve the scaling-fidelity dilemma, we decompose the conditioning signal into a \textbf{(TIB)}. Specifically, we learn a compact representation $z = \mathcal{F}_{\text{pool}}(\mathcal{P})$, where $z \in \mathbb{R}^{K \times d}$ and $K \ll L$. The density of the conditioning information is controlled by the stride parameter $s$, such that $K \approx \lceil L/s \rceil$.

In our implementation, the bottleneck is computed deterministically, and we model
\begin{align}
    p(x \mid \mathcal{P}) = p\big(x \mid z(\mathcal{P})\big).
\end{align}
That is, $p(z \mid \mathcal{P})$ is implemented as a deterministic pooling module over residue embeddings (Sec.~\ref{sec:meth_tib}). The conditional distribution $p(x \mid z)$ is defined by a transformer-based denoiser operating in the SMILES space within the discrete diffusion framework.

\subsection{Topological Information Bottleneck via Soft Assignment Pooling}
\label{sec:meth_tib}
The core of SiDGen's efficiency lies in the learnable pooling mechanism. Unlike fixed downsampling, which can indiscriminately discard important residues in the binding pocket, our prototype-based soft assignment pooling (DiffPool-inspired) learns a soft assignment matrix $\mathbf{S} \in \mathbb{R}^{L \times K}$ that preserves the topological integrity (neighbourhood and connectivity) of the protein active site.

\paragraph{Why this pooling is adaptive and information-preserving.} The assignment weights $\mathbf{S}_{b,i,k}$ are calculated from learned projections of residue tokens and prototypes, so the pooling is \emph{input-dependent}: residues that are most informative for the target (\textit{e.g.}, pocket residues) can be assigned with higher mass to a small subset of bottleneck tokens. Moreover, the pooled token $\mathbf{z}_{b,k}=\sum_i \mathbf{S}_{b,i,k}\mathbf{X}_{b,i}$ is a convex combination of residue embeddings; this preserves information in the sense that each bottleneck token remains in the span of the original residue features while enabling end-to-end differentiability through $\mathbf{S}$.


\paragraph{Encoder Architecture.} 
Given residue embeddings $\mathbf{X} \in \mathbb{R}^{B \times L \times d}$, we first form $K=\lceil L/s \rceil$ pooled prototypes. 
We compute soft assignments using learned projections and scaled dot-product similarity:
\begin{align}
\ell_{b,i,k} &= \frac{\langle \phi_{\text{tok}}(\mathbf{X}_{b,i}),\, \phi_{\text{proto}}(\mathbf{P}_{b,k}) \rangle}{\sqrt{d}}, \\
\mathbf{S}_{b,i,:} &= \text{softmax}(\ell_{b,i,:}).
\end{align}
For each batch element $b$, residue $i$, and prototype $k$, we compute a similarity score $\ell_{b,i,k}$ as the dot product between the projected residue and prototype embeddings, scaled by $\sqrt{d}$. A softmax over $k$ gives the assignment weights, $\mathbf{S}_{b,i,:}$. We mask padded residues and invalid prototypes before the softmax. 

The bottleneck tokens are pooled as weighted aggregation of the residue embeddings,
\begin{align}
\mathbf{z}_{b,k} = \sum_{i=1}^{L} \mathbf{S}_{b,i,k}\,\mathbf{X}_{b,i} \in \mathbb{R}^{d}, \quad \Rightarrow\quad \mathbf{z} \in \mathbb{R}^{B \times K \times d}.
\end{align}
This bottleneck is compact enough to support coarse-grid pairwise computations and efficient cross-attention conditioning in the diffusion model.


\subsection{Chemformer Interface and Discrete Diffusion}
\label{sec:meth_diffusion}
SiDGen utilizes a transformer-based denoiser with a Chemformer-style tokenizer/architecture \citep{irwin2022chemformer} as the denoising network $f_\theta$. The input to the decoder is a masked SMILES sequence $x_t \in \{s_1, \dots, s_M\}$. 

\paragraph{Ligand tokenization and embeddings.} We represent each ligand as a sequence of token IDs produced by a Chemformer-style SMILES tokenizer, including special tokens such as [PAD] and [MASK]. Let $x_t \in \{1,\ldots,V\}^{B\times L_{\text{lig}}}$ denote the (possibly corrupted) token IDs at diffusion time $t$. The denoiser maps token IDs to continuous vectors \textit{via} a learned embedding table $\mathrm{Embed}:\{1,\ldots,V\}\to\mathbb{R}^{H}$, producing initial ligand hidden states
\begin{align}
\mathbf{H}^{(0)} = \mathrm{Embed}(x_t) \in \mathbb{R}^{B\times L_{\text{lig}}\times H}.
\end{align}
These embeddings are then processed by a transformer decoder stack with rotary position embeddings (RoPE) in the ligand self-attention layers.

\paragraph{Topological Conditioning.} The pooled protein representation is injected into every layer of the transformer decoder \textit{via} cross-attention over the conditioning tokens $\mathbf{E}\in\mathbb{R}^{B\times K\times H}$. For a ligand query vector $q_i \in \mathbb{R}^H$, the attention update is:
\begin{align}
    \text{CrossAttn}(q_i, \mathbf{E}) = \text{softmax}\left(\frac{q_i \mathbf{W}_Q (\mathbf{E} \mathbf{W}_K)^\top}{\sqrt{H}}\right) (\mathbf{E} \mathbf{W}_V).
\end{align}
By querying the compressed structural features $\mathbf{E}$, the ligand generator incorporates protein context without attending over all $L_{\text{prot}}$ residues.





\subsection{Decoder Architecture}

The denoiser is a multi-layer Transformer decoder that updates ligand hidden states {via} (i) ligand self-attention (with RoPE) and (ii) cross-attention into a protein-derived conditioning context. In this section, we use $\mathbf{X}_\ell \in \mathbb{R}^{B\times L_{\text{lig}}\times H}$ for the ligand hidden states at decoder layer $\ell$; protein residue embeddings are denoted separately as $\mathbf{X}^{\text{prot}} \in \mathbb{R}^{B\times L_{\text{prot}}\times d}$.

\textbf{Inputs:}
\begin{enumerate}
    \item [(i)] $t \in \mathbb{R}^B$: scalar timesteps for each batch element $b=1,\dots,B$.  
    \item [(ii)] $x_t \in \{1,\ldots,V\}^{B\times L_{\text{lig}}}$: ligand token IDs after corruption at diffusion time $t$.  
    \item [(iii)]$\mathbf{X}_0 = \mathrm{Embed}(x_t) \in \mathbb{R}^{B \times L_{\text{lig}} \times H}$: embedded ligand tokens (decoder input states).  
    \item [(iv)]$\mathbf{E} \in \mathbb{R}^{B \times K \times H}$: bottleneck conditioning tokens of length $K$ (typically $K=\lceil L_{\text{prot}}/s \rceil$), obtained by pooling protein features and projecting to hidden size $H$.  
\end{enumerate}

\textbf{Timestep Embeddings:} Sinusoidal encodings are projected through a two-layer MLP with SiLU \cite{elfwing2017sigmoidweightedlinearunitsneural} activations to form the timestep token:  
\begin{align}
\sigma(t) = \text{SiLU}\!\big(W_2 \,\text{SiLU}(W_1 \,\text{PE}(t))\big), \quad \sigma(t) \in \mathbb{R}^H.
\end{align}

\textbf{Conditioning context:} The diffusion timestep is embedded as a single token and prepended to the conditioning tokens to form the cross-attention key/value sequence
\begin{align}
\mathbf{C}_{\text{cond}} = [\,\sigma(t);\; \mathbf{E};\; \mathbf{X}^{\text{prot}}\,].
\end{align}

\textbf{Rotary Position Embeddings (RoPE)~\citep{su2024roformer}:}
\begin{align}
\text{RoPE}(x_m, m) &= R(m) x_m, \\
R(m) &= 
\begin{pmatrix}
\cos(m\theta) & -\sin(m\theta)\\
\sin(m\theta) & \cos(m\theta)
\end{pmatrix}, \\
\theta &= 10000^{-2i/d},
\end{align}
where $x_m$ is the $m$-th token embedding, $m$ is the position index, and $d=H/h$ is the per-head dimension.

\paragraph{Multi-Head Attention Mechanism}
Each decoder layer $\ell$ applies the following operations sequentially:

\textbf{(i) Self-Attention on Ligand Tokens:} The ligand sequence attends to itself to capture intra-molecular dependencies:
\begin{align}
Q_{\text{lig}} &= \mathbf{X}_{\ell-1} W_Q^{\text{self}}, \\
K_{\text{lig}} &= \mathbf{X}_{\ell-1} W_K^{\text{self}}, \quad V_{\text{lig}} = \mathbf{X}_{\ell-1} W_V^{\text{self}}, \\
\mathbf{A}_{\text{self}} &= \text{Softmax}\left(\frac{Q_{\text{lig}} K_{\text{lig}}^T}{\sqrt{d}}\right) V_{\text{lig}}, \\
\mathbf{X}'_{\ell} &= \text{LayerNorm}(\mathbf{X}_{\ell-1} + \mathbf{A}_{\text{self}}).
\end{align}
Here, the ligand tokens attend to each other to capture intra-SMILES dependencies; timestep information enters through cross-attention to $\mathbf{C}_{\text{cond}}$.

\textbf{(ii) Cross-Attention with Conditioning Context:} The ligand tokens query the conditioning context to incorporate target information:
\begin{align}
Q_{\text{lig}} &= \mathbf{X}'_{\ell} W_Q^{\text{cross}}, \\
K_{\text{cond}} &= \mathbf{C}_{\text{cond}} W_K^{\text{cross}}, \quad V_{\text{cond}} = \mathbf{C}_{\text{cond}} W_V^{\text{cross}}, \\
\mathbf{A}_{\text{cross}} &= \text{Softmax}\left(\frac{Q_{\text{lig}} K_{\text{cond}}^T}{\sqrt{d}}\right) V_{\text{cond}}, \\
\mathbf{X}''_{\ell} &= \text{LayerNorm}(\mathbf{X}'_{\ell} + \mathbf{A}_{\text{cross}}).
\end{align}
This operation allows each ligand token to attend to all conditioning tokens, aggregating relevant target information. In our implementation, $\mathbf{C}_{\text{cond}}$ is formed from the timestep token and bottleneck tokens (and the full residue sequence embeddings from ESM-2).

\textbf{(iii) Feed-Forward Network:} Position-wise MLP with residual connection:
\begin{align}
\text{FFN}(x) &= W_2 \, \text{ReLU}(W_1 x + b_1)  + b_2, \\
\mathbf{X}_{\ell} &= \text{LayerNorm}(\mathbf{X}''_{\ell} + \text{FFN}(\mathbf{X}''_{\ell})).
\end{align}

The final layer output is projected to vocabulary logits for token prediction.

\subsection{Coarse Folding of Structural Features on the Bottleneck Grid}
\label{sec:folding}

\paragraph{Coarsening with stride $s$.}
Before learned pooling, we define a simple coarsening operator that subsamples residues at stride $s$ to form a coarse grid of size $K=\lceil L/s \rceil$.
Let $\mathcal{I}_s = \{0, s, 2s, \dots\}$ be the strided index set. For single and pair features,
\begin{align}
\mathbf{s}^{(s)} &= \mathbf{s}[\mathcal{I}_s] \in \mathbb{R}^{B \times K \times C_{\text{single}}}, \\
\mathbf{p}^{(s)} &= \mathbf{p}[\mathcal{I}_s,\mathcal{I}_s] \in \mathbb{R}^{B \times K \times K \times C_{\text{pair}}}.
\end{align}
This fixed coarsening reduces memory from $O(L^2)$ to $O(K^2)$, and serves as the baseline we compare against our learned bottleneck.

To mitigate the quadratic scaling of structural features, we employ a learnable soft assignment mechanism inspired by DiffPool~\citep{ying2018hierarchical}. Given protein sequence embeddings \(\mathbf{X} \in \mathbb{R}^{B \times L \times d}\), we learn an assignment matrix that pools residue embeddings to a coarser resolution.

\paragraph{Soft Assignment Matrix} We compute \(\mathbf{S} \in \mathbb{R}^{B \times L \times K}\) using strided prototypes and learned projections (Sec.~\ref{sec:meth_tib}). This yields a lightweight soft clustering while keeping computation efficient.

\paragraph{Coarse Pair Construction} We pool residue embeddings into bottleneck tokens \(\mathbf{z} \in \mathbb{R}^{B \times K \times d}\), and then construct coarse pair features directly on the bottleneck grid:
\begin{align}
\mathbf{p}_c = \text{OuterLinearPair}(\mathbf{z}) \in \mathbb{R}^{B \times K \times K \times C_{\text{pair}}}.
\end{align}
This avoids forming any full-resolution \(L\times L\) pair tensor and yields the claimed \(O(K^2)\) pairwise footprint.

\paragraph{Folding Operations} On the coarse pair features $\mathbf{p}_c \in \mathbb{R}^{B \times K \times K \times C_{\text{pair}}}$, we apply triangle attention and triangle multiplication inspired by AlphaFold ~\cite{jumper2021highly}:

\textbf{Triangle Attention:}
\begin{align}
\text{TriAttn}_{\text{start}}(\mathbf{p}_c):\quad \mathbf{p}_c[b,i,j] &\leftarrow \text{Attention}_k(\mathbf{p}_c[b,i,k]), \\
\text{TriAttn}_{\text{end}}(\mathbf{p}_c):\quad \mathbf{p}_c[b,i,j] &\leftarrow \text{Attention}_k(\mathbf{p}_c[b,k,j]).
\end{align}

\textbf{Triangle Multiplication:}
\begin{align}
\text{TriMult}_{\text{out}}(\mathbf{p}_c):\; \mathbf{p}_c[b,i,j] &\leftarrow \sum_k \mathbf{p}_c[b,i,k] \odot \mathbf{p}_c[b,j,k], \\
\text{TriMult}_{\text{in}}(\mathbf{p}_c):\; \mathbf{p}_c[b,i,j] &\leftarrow \sum_k \mathbf{p}_c[b,k,i] \odot \mathbf{p}_c[b,k,j].
\end{align}

These operations process the coarse structure, updating pairwise relationships among the \(K\) bottleneck tokens. Triangle attention is quadratic in \(K\), while triangle multiplication additionally contracts over \(K\) (\(O(K^3)\)).

\paragraph{Decoder Conditioning} The decoder consumes the processed bottleneck tokens as conditioning. We derive a coarse cross-attention key-bias per head from aggregated coarse pair features.

\paragraph{Complexity Analysis} The bottleneck-based mechanism reduces memory and compute while enabling learned feature aggregation:
\begin{itemize}
    \item \textbf{Memory:} Pair features scale as \(O(L^2) \rightarrow O(K^2)\) where \(K = \lceil L/s \rceil\). For \(s=4\), this yields \(\sim16\times\) reduction.
    \item \textbf{Compute:} Triangle attention/multiplication operations scale as \(O(K^2 \cdot K) = O(K^3)\) vs. \(O(L^3)\), providing \(\sim s^3 = 64\times\) speedup for \(s=4\).
    \item \textbf{Assignment overhead:} Computing \(\mathbf{S}\) requires \(O(L \cdot K)\) operations, which is negligible compared to the \(O(L^2)\) savings.
    \item \textbf{Gradient flow:} Unlike fixed downsampling, soft assignments enable end-to-end learning, allowing the model to adapt pooling to task-specific structural patterns.
\end{itemize}


\subsection{Training Enhancements}

\textbf{Substitution Parameterization:} masked diffusion loss with substitution scaling:
\begin{align}
\mathcal{L_\text{MDLM}} = -\sum_{i,t} \log p_\theta(x_0^{(i)} \mid x_t^{(i)}, c) \cdot \frac{d\sigma/dt}{\exp(\sigma)-1}.
\end{align}

\textbf{Masking corruption:} we sample $t\sim\mathrm{Uniform}(0,1)$, compute a noise level $\sigma(t)$, and independently replace each token with [MASK] with probability $1-\exp(-\sigma(t))$.

\paragraph{Curriculum Learning}  
To gradually increase the difficulty of the denoising task, we scale the timestep used for loss computation:
\begin{align}
t_{\text{curriculum}} &= \min\big(\max(\epsilon, \alpha_{\text{epoch}} \, t), 1\big), \\
\alpha_{\text{epoch}} &= \min\Big(1, \frac{\text{epoch}+1}{T_{\text{curriculum}}}\Big),
\end{align}
where \(t \in [0,1]\) is the diffusion timestep, \(\epsilon>0\) is a small lower bound, \(\text{epoch}\) is the current training epoch, and \(T_{\text{curriculum}}\) is the total number of curriculum epochs.  


\paragraph{In-loop Validity Penalty.} We define a sample-based validity regularizer that directly targets the model's empirical valid-generation rate. Formally the desideratum is
\begin{align}
    \mathcal{J}_{\text{valid}} \;=\; \mathbb{E}_{x\sim p_\theta}\big[\mathbf{1}_{x\ \text{valid}}\big]
\end{align}

the expected fraction of valid SMILES under the model distribution. In practice, we approximate this expectation by Monte--Carlo: every $N$ optimizer steps we decode $S$ samples from the current model and compute the empirical validity
\begin{align}
     \widehat{\mathcal{J}}_{\text{valid}} \;=\; \frac{1}{S}\sum_{i=1}^S \mathbf{1}_{x_i\ \text{valid}},
\end{align}
and optimize it using a score-function (REINFORCE) estimator. Because decoding and the validity indicator are non-differentiable, we do not backpropagate through the sampling/validation pipeline; instead, we attach gradients through the log-probability of the sampled sequence:
\begin{align}
\nabla_\theta\,\mathbb{E}_{x\sim p_\theta}[\mathbf{1}_{x\ \text{valid}}]
\;=\;
\mathbb{E}_{x\sim p_\theta}\big[\mathbf{1}_{x\ \text{valid}}\,\nabla_\theta \log p_\theta(x)\big].
\end{align}
Concretely, every $N$ optimizer steps we sample $S$ sequences, compute validity flags, and add the REINFORCE loss
\begin{align}
\mathcal{L}_{\text{total}} = \mathcal{L}_{\text{MDLM}} - \lambda_{\text{valid}}\,\frac{1}{S}\sum_{i=1}^S \big(\mathbf{1}_{x_i\ \text{valid}}-b\big)\,\log p_\theta(x_i),
\end{align}
where $b$ is a moving baseline (implemented as the batch mean validity) to reduce gradient variance.


\section{Experiments}
\label{sec:experiments}

\subsection{Evaluation Framework and Datasets}
\label{sec:exp_datasets}
We evaluate SiDGen across four complementary benchmarks to assess molecular quality, structural fidelity, and virtual screening performance.
\begin{itemize}
    \item \textbf{MOSES} \citep{10.3389/fphar.2020.565644}: Used for assessing unconditional molecular generation quality, including chemical validity, uniqueness, and novelty.
    \item \textbf{CrossDocked2020} \citep{francoeur2020three}: A large-scale protein-ligand dataset used to evaluate 3D binding affinity and pose reasonability.
    \item \textbf{DUD-E} \citep{mysinger2012directory}: Evaluates high-throughput virtual screening capabilities using enrichment factors (EF@1\%), FCD \cite{preuer2018frechetchemnetdistancemetric} and BEDROC \cite{Zhao2009} scores.
\end{itemize}

\subsection{Molecular Quality on MOSES}
\label{sec:exp_moses}
Table \ref{tab:moses} summarizes SiDGen's performance on the MOSES benchmark. Our Chemformer-style transformer denoiser achieves 100\% validity and competitive novelty (100\%), significantly outperforming previous VAE and GAN-based baselines. The high uniqueness (88.75\%) confirms that SiDGen does not simply memorize the training set but explores the latent chemical space efficiently.

\begin{table}[h]
\centering
\caption{Molecular generation quality on the MOSES benchmark.}
\label{tab:moses}
\begin{tabular}{lccccc}
\toprule
Method & Validity & Uniqueness & Novelty & IntDiv $\uparrow$ \\
\midrule
CharRNN & 97.5\% & 99.9\% & 84.2\% & 0.856 \\
VAE & 97.7\% & 99.8\% & 69.5\% & 0.855 \\
JTN-VAE & 100\% & 99.9\% & 91.4\% & 0.855 \\
\midrule
\textbf{SiDGen} & \textbf{100\%} & \textbf{88.75\%} & \textbf{100\%} & \textbf{0.903} \\
\bottomrule
\end{tabular}
\end{table}

\subsection{Structure-Based Benchmarks: CrossDocked2020}
\label{sec:exp_crossdock}
We evaluate SiDGen's ability to generate target-specific ligands on the CrossDocked2020 test set. We compare SiDGen against both classical 3D generators and recent 2024-2025 SOTA models.

\begin{table*}[t]
\centering
\caption{Performance comparison on the CrossDocked2020 test set.}
\label{tab:crossdock_full}
\begin{tabular}{lcccccc}
\toprule
Model & Paradigm & Vina Dock $\downarrow$ & QED $\uparrow$ & SA $\uparrow$ & Diversity $\uparrow$ & Time(s) $\downarrow $ \\
\midrule
\citelink{peng2022pocket}{Pocket2Mol} & 3D Autoregressive & -7.25 & 0.56 & {0.76} & 0.86 & 2184 \\
\citelink{guan2023targetdiff}{TargetDiff} & 3D Diffusion & -7.46 & 0.48 & 0.58 & 0.72 & 3156\\
\citelink{guan2023decompdiff}{DecompDiff} & 3D Decomposition & -8.39 & 0.45 & 0.61 & 0.87  & 5367\\
\citelink{wang2024molcraft}{MolCRAFT} & Fragment-based & -9.25 & 0.46 & 0.58 & 0.82 & 139 \\
\citelink{Mistryukova2025.06.16.659955}{ProtoBindDiff} & Diffusion Based &-9.03 & 0.55 &  0.33 & 0.88 & 143 \\
\citelink{powers2025rxnflow}{RxnFlow} & Reaction-aware & -8.85 & \textbf{0.67} & 0.35 & 0.81 & \textbf{4}\\
\citelink{chen2025cidd}{CIDD} & LLM-Guided 3D & -9.02 & {0.53} & 0.69 & 0.87 & 149\\
\citelink{qiu2025piloting}{MolJO} &  Fragment Based Optimizer& -9.05 & 0.56 &  \textbf{0.78} & 0.66 & 18\\
\midrule
\textbf{SiDGen (Ours)} & \textbf{HSD + TIB} & \textbf{-9.81} & 0.58 & 0.49 & \textbf{0.90} & 63 \\
\bottomrule
\end{tabular}
\end{table*}

As shown in Table \ref{tab:crossdock_full}, SiDGen achieves a superior Vina Dock score of -9.81 kcal/mol, outperforming even recent physics-aware 3D generators. Crucially, SiDGen maintains significantly higher diversity (0.89), suggesting that the Topological Information Bottleneck allows for broader exploration of valid binding modes compared to rigid 3D coordinate-based sampling.

\subsection{Virtual Screening Performance on DUD-E}
\label{sec:exp_dude}
To evaluate practical utility in drug discovery pipelines, we conduct a virtual screening study on the DUD-E dataset. We measure the model's ability to rank active compounds above decoys.

\begin{table}[h]
\centering
\caption{DUD-E Virtual Screening metrics. EF@1\% measures early enrichment, while BEDROC assesses ranking quality.}
\label{tab:dude_benchmarks}
\begin{tabular}{lcccc}
\toprule
Method & ROC-AUC & EF@1\% & BEDROC \\
\midrule
\citelink{doi:10.1021/acs.jcim.1c00203}{AutoDock Vina} & 0.697 & 8.82 & 0.15 \\
\citelink{McNutt_Francoeur_Aggarwal_Masuda_Meli_Ragoza_Sunseri_Koes_2021}{Gnina} & 0.680 & 7.93 & 0.18 \\
\citelink{doi:10.1021/jm051256o}{Glide SP} & 0.830 & 9.5 & 0.29 \\
\midrule
\textbf{SiDGen} & \textbf{0.819} & \textbf{10.26} & \textbf{0.28} \\
\bottomrule
\end{tabular}
\end{table}

The competitive enrichment factor (EF@1\% = 10.26) demonstrates that SiDGen's structural conditioning is sufficient to distinguish subtle differences in binding complementarity, even without explicit 3D coordinate generation at inference time.
\subsection{Training Components}
Table~\ref{tab:abl_training_full} ablates curriculum learning and validity regularization on CrossDocked2020 benchmark.

\begin{table}[t]
\centering
\small
\setlength{\tabcolsep}{6pt}
\caption{Ablation of training enhancements on CrossDocked2020 }
\label{tab:abl_training_full}
\begin{tabular}{ccc|c|c}
\toprule
Folding  & Curriculum & Val Penalty & Vina Dock & QED\\
\midrule
\xmark & \xmark & \xmark & -9.03 & 0.55 \\
 \cmark & \xmark & \xmark &  -9.26 & 0.53\\
 \cmark & \cmark & \xmark & -9.66 & 0.54\\
\cmark & \xmark & \cmark & -9.57 & 0.56\\
\midrule
 \cmark & \cmark & \cmark & \textbf{-9.74} & \textbf{0.58} \\
\bottomrule
\end{tabular}
\end{table}

Both components contribute synergistically: curriculum learning and validity regularization eliminate invalid outputs. Combined, they achieve perfect validity while getting better quality results than baseline.
\subsection{Resolving the Scaling-Fidelity Dilemma}
\label{sec:exp_scaling}

\begin{figure}[h]
    \centering
    \fbox{
        \includegraphics[width=0.9\linewidth]{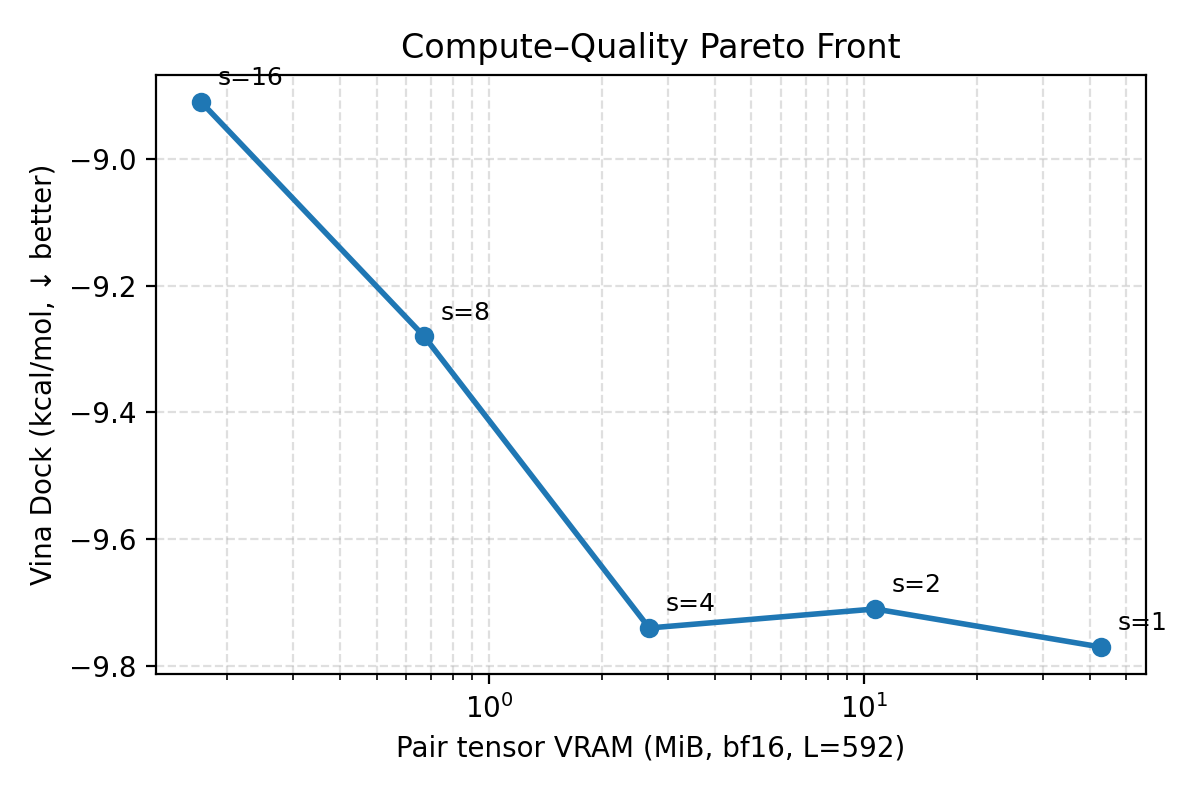}
    }
    \caption{The Compute-Quality Pareto Front.}
    \label{fig:pareto}
\end{figure}

\paragraph{Pareto Analysis: Computation vs. Quality.}
We systematically vary the pooling stride $s$ from 1 (dense) to 16 (ultra-coarse) to investigate the compute-quality trade-off. As shown in Figure \ref{fig:pareto}, increasing the stride significantly reduces VRAM usage while maintaining robust binding scores. 
We additionally compare the learned soft-assignment bottleneck to a coarse strided pooling baseline (index-select downsampling in the folding block); learned pooling consistently yields better binding and fidelity at the same stride. We report this comparison in Appendix~\ref{app:ablations} (Table~\ref{tab:abl_pooling}).

\paragraph{Pocket Fidelity Analysis.}
To quantify information loss induced by the bottleneck, we compute a \emph{pocket-fidelity recovery} score on residues proximal to the bound ligand. For each protein--ligand complex, let $\mathcal{I}_{\text{pocket}}$ be the set of residues with minimum heavy-atom distance $\leq 5\,\text{\AA}$ to any ligand heavy atom. We partition $\mathcal{I}_{\text{pocket}}$ into (i) hydrophobic-contact residues (hydrophobic sidechains within $5\,\text{\AA}$) and (ii) hydrogen-bond-capable residues (donor/acceptor sidechains within $5\,\text{\AA}$). For a given pooling stride $s$, we compute the bottleneck tokens $\mathbf{z}$ and reconstruct residue-level embeddings as $\widehat{\mathbf{X}} = \mathbf{S}\mathbf{z}$.

We define the recovery score for a residue subset $\mathcal{I}$ as the mean cosine similarity between original and reconstructed embeddings:
\begin{align}
\mathrm{Rec}(\mathcal{I}) = 100\cdot \frac{1}{|\mathcal{I}|}\sum_{i\in\mathcal{I}} \frac{\langle \mathbf{X}_{i},\widehat{\mathbf{X}}_{i}\rangle}{\|\mathbf{X}_{i}\|\,\|\widehat{\mathbf{X}}_{i}\|}.
\end{align}
This yields an interpretable percentage of how well pocket-local information is preserved by the bottleneck; learnable pooling attains 89.2\% recovery at $s=4$, whereas fixed-stride downsampling attains 61.1\%.

\begin{figure}[h]
    \centering
    \fbox{
        \includegraphics[width=0.9\linewidth]{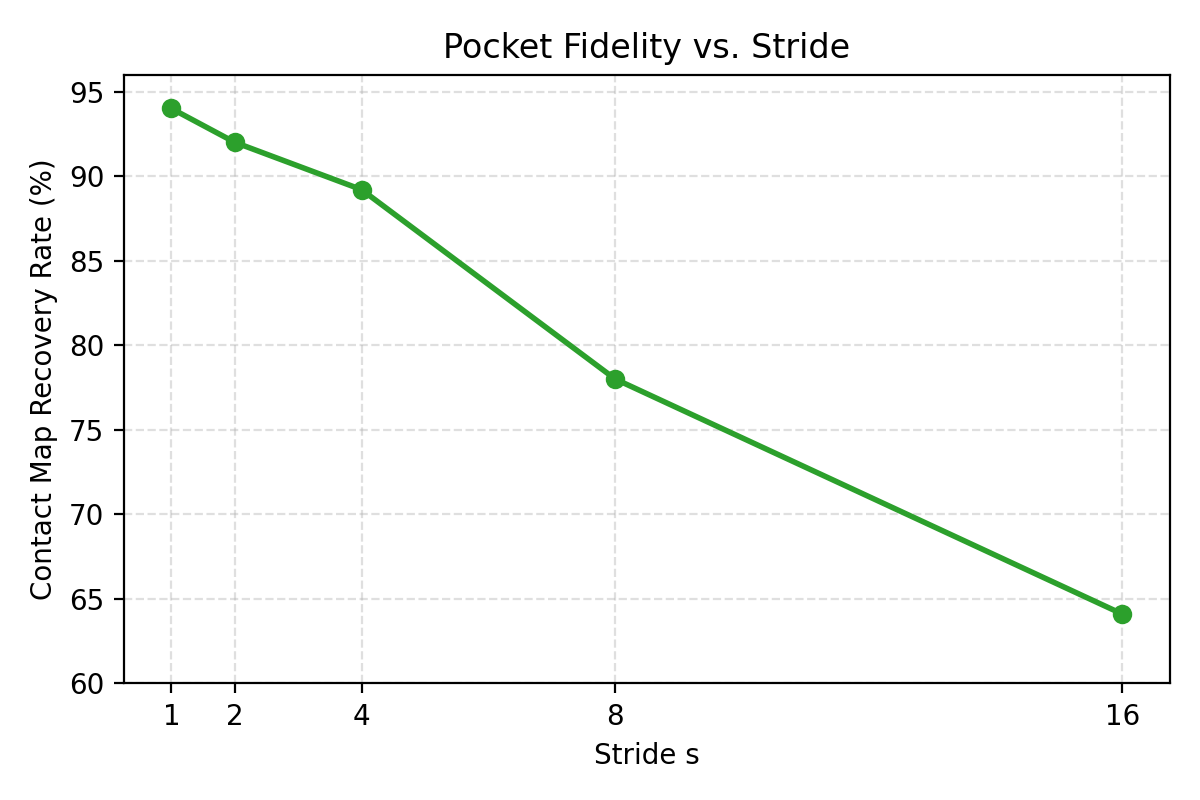}
    }
    \caption{Pocket Fidelity Quantification: Percent recovery of hydrophobic and hydrogen-bond interactions within 5\AA of the ligand.}
    \label{fig:fidelity}
\end{figure}

\section{Discussion}
\label{sec:discussion}

\subsection{Innovation: The Topological Perspective}
\label{sec:disc_innovation}
One common critique of string-based generative models for SBDD is their perceived lack of geometric awareness. However, we argue that the \textbf{Topological Information Bottleneck} (TIB) represents a shift from "physical modeling" to "topological abstraction." While 3D coordinate generators attempt to satisfy physical constraints (e.g., steric clashing) through explicit force fields or E(3)-equivariant updates, SiDGen learns to satisfy these constraints through latent topological descriptors. 

The success of the TIB suggests that the chemical space of binding pockets is lower-dimensional than the raw coordinate space. By compressing $L$ residues into $K$ topological clusters, SiDGen focuses on the \emph{shape} and \emph{complementarity} of the interaction surface rather than the precise position of every atom. This mirrors how expert medicinal chemists often reason about the "key interactions" (e.g., a specific salt bridge or hydrophobic pocket) rather than the entire protein volume.

\subsection{Efficiency and High-Throughput Readiness}
\label{sec:disc_efficiency}
The reduction of memory scaling from $O(L^2)$ to $O(K^2)$ enables two major use cases. First, it allows for the generation of ligands for large, multi-subunit protein complexes (e.g., ribosome-bound inhibitors) that are currently beyond the reach of dense structural generators. Second, it enables \textbf{generative virtual screening}, where the model can be conditioned on thousands of different pocket conformations (from MD simulations) in a fraction of the time required for physics-based docking.

\subsection{Synthesizability and Reaction-Awareness}
\label{sec:disc_synth}
While SiDGen achieves 100\% validity and high diversity, the generation of "drug-like" molecules also requires consideration of synthesizability. Unlike 3D generators that may produce strained geometries, SiDGen's SMILES-space transformer denoiser (Chemformer-style architecture) encourages outputs that remain consistent with the known chemical manifold of druglike SMILES. Future iterations will investigate the integration of reaction-aware tokens, as proposed in \textbf{RxnFlow} \citep{powers2025rxnflow}, to further improve the down-stream utility of the generated hits.

\subsection{Limitations and Future Work}
\label{sec:disc_limitations}
The primary limitation of SiDGen is the absence of an explicit 3D pose for the generated ligand. Although the Vina Dock scores are high, these require an external docking step at inference. We are currently developing a "joint-diffusion" variant in which the TIB conditions both the SMILES generation and a coarse-grained 3D pose prediction. This would allow for end-to-end "pose-aware" generation without sacrificing the efficiency of the topological bottleneck.

\section{Conclusion}
\label{sec:conclusion}
We have presented \textbf{SiDGen}, a hierarchical structural diffusion model that resolves the scaling-fidelity dilemma in SBDD. By introducing the Topological Information Bottleneck via learned soft assignment pooling (DiffPool-inspired), we achieve a significant reduction in pairwise-tensor complexity while maintaining state-of-the-art binding affinities and virtual screening performance. SiDGen demonstrates that efficient, pocket-aware molecular generation is achievable through structured topological compression, providing a scalable and expressive framework for the next generation of AI-driven drug discovery.

\section*{Impact Statement}
This paper is aimed to facilitate in-silico rational drug design.
Potential society consequences include mal-intended usage
of toxic compound discovery, which needs support from
professional wet labs and thus expensive to reach. Therefore
we do not possess a negative vision that this might lead to
serious ethical consequences, though we are aware of such
a possibility.
\bibliography{example_paper}
\bibliographystyle{icml2026}

\clearpage
\appendix
\onecolumn
\section{Detailed Experimental Configuration}
\label{app:exp_details}

\subsection{Data Preprocessing}
\begin{itemize}
\item \textbf{SMILES Canonicalization:} SMILES are tokenized with a Chemformer-style SMILES tokenizer (the JSON tokenizer used in the repo); on-the-fly SMILES randomization is enabled.
\item \textbf{Sequence Processing:} We use precomputed ESM-2 embeddings loaded; sequences are padded per-batch using length-aware batching.
\item \textbf{Metadata Files:} The diffusion dataset index (data.csv) contains 1,154,055 entries, and categorical mappings include 793,429 entries; the tokenizer vocabulary size is 54 tokens.
\item \textbf{Data Splitting:} Random train/validation/test split of 80/10/10 (923,244 / 115,406 / 115,405 samples; total 1,154,055) with seed 188.
\end{itemize}

\subsection{Model Architecture Details}
\begin{itemize}
\item \textbf{Transformer Configuration:} 4 decoder layers, hidden dimension $H=1280$, 8 attention heads ($d_{\text{head}}=160$), FFN intermediate dimension $3H=3840$
\item \textbf{Pooling/Folding Parameters:} Stride $s=4$, pair dimension $C_{\text{pair}}=64$, folding heads 8, folding head dim 16, transition factor 3.
\item \textbf{Vocabulary:} SMILES tokenizer with 54 tokens including standard SMILES characters (C, N, O, etc.), brackets, ring numbers, special tokens ([MASK], [PAD], [BOS], [EOS])
\item \textbf{Maximum Lengths:} Protein $L_{\text{max}}=1024$ (evaluation), SMILES $L_{\text{SMILES}}=1700$ (training default)
\item \textbf{Position Encodings:} RoPE (base 10000) applied in ligand self-attention; cross-attention uses standard dot-product attention.
\item \textbf{Layer Normalization:} Decoder uses post-norm residual blocks (Add $\rightarrow$ LayerNorm); folding modules use LayerNorm without affine parameters.
\end{itemize}

\subsection{Training Details}
\begin{itemize}
\item \textbf{Optimizer:} AdamW with $\beta_1=0.9$, $\beta_2=0.999$, $\epsilon=10^{-8}$, weight decay $0.01$
\item \textbf{Learning Rate Schedule:} Constant learning rate $10^{-4}$ (AdamW) for all training steps.
\item \textbf{Batch Configuration:} Max per-device batch size 16, batch volume 8,000,000, gradient accumulation steps 2
\item \textbf{Gradient Clipping:} Max norm 1.0 applied to global gradient norm
\item \textbf{Curriculum Learning:} $T_{\text{curriculum}}=2$ epochs, linearly increasing $\alpha_{\text{epoch}}$ from 0.1 to 1.0
\item \textbf{Validity Penalty:} $\lambda_{\text{valid}}=0.1$, computed on 32 mini-batch samples every 100 steps using RDKit SMILES validation
\item \textbf{Training Duration:} 10 epochs on 1,154,055 samples (steps depend on effective batch size)
\item \textbf{Precision:} bf16 mixed precision (float type \texttt{bfloat16})
\item \textbf{Hardware:} A100-class GPUs; peak per-GPU VRAM $\approx$60 GB (including batch + backprop)
\item \textbf{Checkpointing:} Model checkpoints saved every \texttt{checkpoint\_every=10000} optimizer steps; best model selected based on validation SMILES validity.
\end{itemize}

\subsection{Inference Configuration}
\begin{itemize}
\item \textbf{Sampling Steps:} $T=100$ diffusion steps with log-linear noise schedule
\item \textbf{Temperature:} 1.0 (no temperature scaling for sampling)
\item \textbf{Top-k/Top-p:} Nucleus sampling with $p=0.9$; no top-k
\item \textbf{Noise Parameters:} $\eta=0.1$, noise removal enabled
\item \textbf{Guidance:} Classifier-free guidance disabled ($w=0$, \texttt{cfg\_every\_k=0})
\item \textbf{Generation Count:} 100 molecules per target protein for all evaluations
\item \textbf{Post-processing:} Generated SMILES are canonicalized and validated with RDKit. 

\end{itemize}

\subsection{Evaluation Protocol Details}
\begin{itemize}
\item \textbf{Docking Setup:} We use a fixed docking engine (AutoDock Vina) with a single standardized protocol across baselines: identical search box sizes, exhaustiveness, and number of poses retained. We report Vina Dock (kcal/mol) on the top-ranked pose.
\item \textbf{Success Rate:} A generated pose is counted as a success if it achieves Vina Dock $\leq$ a target-specific threshold and is within 2.0\AA\ RMSD of the redocked conformation (when a reference ligand is available).
\item \textbf{Synthesizability:} We report SA score \citep{Ertl2009} and QED \cite{bickerton2012quantifying} for all generated molecules, matching evaluation protocols in recent SBDD benchmarks.
\end{itemize}

\subsection{Molecular Property Distributions}
\begin{figure}[h]
    \centering
    \includegraphics[width=0.8\linewidth]{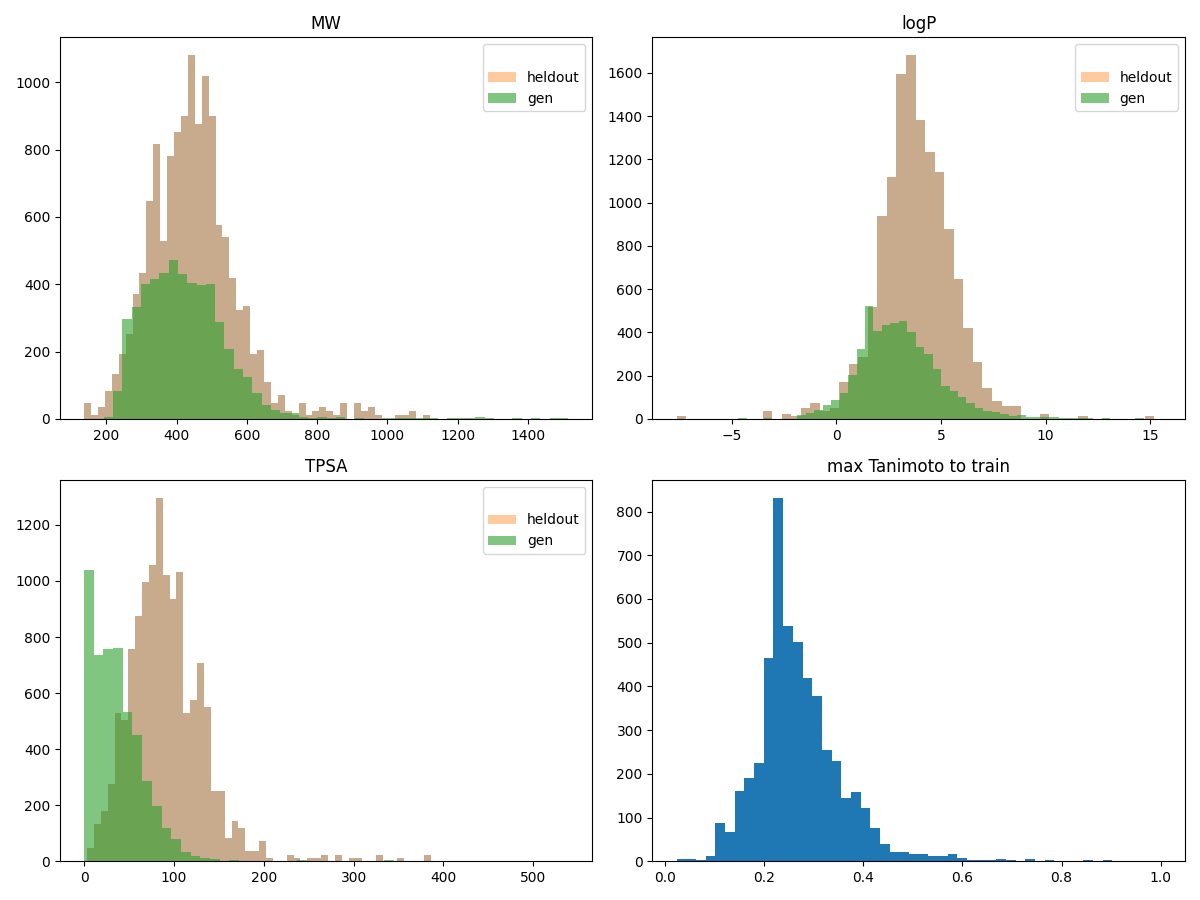}
    \caption{Distributions of key molecular properties }
    \label{fig:prop_dists}
\end{figure}

This figure compares generated (gen) vs holdout distributions for molecular weight (MW), logP, TPSA, and maximum Tanimoto similarity to the training set. The shown run reports mean MW 421.57, mean logP 3.07, mean TPSA 37.89, mean max-Tanimoto 0.27, and SA score 0.49.

\section{Complete Ablation Studies}
\label{app:ablations}

\subsection{Bottleneck Coarse Dimension}
Table~\ref{tab:abl_diffpool_full} shows comprehensive performance vs. coarse dimension $K$ for fixed protein length $L=592$ during inference with $batch\_size = 1$.

\begin{table}[h]
\centering
\caption{Effect of bottleneck coarse dimension $K$ on CrossDocked2020 performance (full results)}
\label{tab:abl_diffpool_full}
\begin{tabular}{lcccccc}
\toprule
Stride $s$ & VRAM (GiB) & Vina Dock & QED & Diversity \\
\midrule
16 & 0.17 & -8.91 & 0.52 & 0.89 \\
8 & 0.67 & -9.28 & 0.54 & 0.88 \\
4 & 2.67 & \textbf{-9.81} & \textbf{0.58} & \textbf{0.90} \\
2 & 10.70 & -9.83 & 0.58 & 0.89\\
1 & 42.78 & -9.87 & 0.59 & 0.91 \\
\bottomrule
\end{tabular}
\end{table}

$K=128$ ($s=4$) provides the best performance-efficiency trade-off. Pair-tensor VRAM drops from 42.78 GiB ($s=1$) to 2.67 GiB ($s=4$) at $L=592$, while preserving near-optimal binding. Very coarse pooling ($K=37$, stride 16) shows 0.96 kcal/mol degradation, confirming the need for sufficient spatial resolution.

\subsection{Pooling Ablation: Learned Pooling vs. Coarse Striding}
We compare learned soft-assignment pooling against a coarse strided pooling baseline (index-select downsampling) at the same stride $s$.

\begin{table}[h]
\centering
\caption{Pooling ablation at fixed stride $s=4$ (CrossDocked2020).}
\label{tab:abl_pooling}
\begin{tabular}{lcccc}
\toprule
Pooling & Vina Dock & QED & SA  \\
\midrule
Coarse Striding & -9.74 & 0.55 & 0.47  \\
Learned Pooling  & -9.81 & 0.58 & 0.49  \\
\bottomrule
\end{tabular}
\end{table}

\section{Cross-Family Performance Analysis}
\label{app:family_analysis}

Table~\ref{tab:family_perf_full} shows detailed performance breakdown by PFAM family on CrossDocked2020 test set.

\begin{table}[h]
\centering
\caption{Performance by protein family (top 5 families by sample count in test set)}
\label{tab:family_perf_full}
\begin{tabular}{lccc}
\toprule
PFAM Family & Count & Vina Dock & QED  \\
\midrule
Kinase (PF00069) & 1247 & -9.89 & 0.56  \\
GPCR (PF00001) & 892 & -9.21 & 0.54  \\
Protease (PF00082) & 634 & -10.11 & 0.57  \\
Nuclear Receptor (PF00104) & 421 & -9.43 & 0.55  \\
Ion Channel (PF00520) & 318 & -8.97 & 0.52  \\
\midrule
Overall & 3512 & -9.81 & 0.58  \\
\bottomrule
\end{tabular}
\end{table}

Performance is consistent across major drug target families. Proteases show the strongest binding affinity (-10.11 kcal/mol), likely due to well-defined active sites. GPCRs and ion channels show slightly lower scores, potentially due to more flexible binding pockets. The 100\% validity across all families demonstrates robustness.

\section{Architectural Details}
\label{app:math_algos}

\subsection{Latent-variable View of the Bottleneck }
\label{app:latent_view}
Conceptually, protein-conditioned generation can be written as a latent-variable model
\begin{align}
    p(x \mid \mathcal{P}) = \int_{\mathcal{Z}} p(x \mid z)\, p(z \mid \mathcal{P})\, dz.
\end{align}
In \textbf{SiDGen}'s implementation, the bottleneck is computed via pooling, i.e., $z = z(\mathcal{P})$. This corresponds to a degenerate posterior $p(z\mid\mathcal{P}) = \delta\big(z - z(\mathcal{P})\big)$, yielding
\begin{align}
    p(x \mid \mathcal{P}) = p\big(x \mid z(\mathcal{P})\big),
\end{align}
which is the view used throughout the paper's main sections and matches the training/inference codepath.

\subsection{Decoder Architecture}
\label{app:decoder_math}
The denoiser is a Transformer decoder operating on masked SMILES token embeddings, with cross-attention into a conditioning context built from a timestep token and bottleneck tokens $z$ and additional protein tokens. The implementation uses post-norm residual blocks (Add $\rightarrow$ LayerNorm).

\paragraph{Timestep token.} We embed timesteps $t\in\mathbb{R}^B$ using sinusoidal features $\text{PE}(t)$ followed by a 2-layer MLP:
\begin{align}
    \tau(t) = \text{SiLU}\!\big(W_2 \,\text{SiLU}(W_1 \,\text{PE}(t))\big) \in \mathbb{R}^{H}.
\end{align}

\paragraph{Sequences.} Let $\mathbf{X}_{\text{lig}}\in\mathbb{R}^{B\times L\times H}$ be ligand token embeddings and $\mathbf{Z}\in\mathbb{R}^{B\times K\times H}$ be the bottleneck tokens. The cross-attention context is
\begin{align}
    \mathbf{C} = [\,\tau(t);\;\mathbf{Z}\; (\mathbf{X}_{\text{prot}})\,].
\end{align}
In the implemented model, $\tau(t)$ participates in cross-attention as part of $\mathbf{C}$ and is not appended to the ligand sequence for self-attention.

\paragraph{RoPE.} Rotary position embeddings (RoPE) are applied to ligand self-attention queries/keys with base frequency 10000; cross-attention uses standard dot-product attention.

\paragraph{One decoder block.} For layer $\ell$, with post-norm residual connections:
\begin{align}
    \mathbf{X}'_{\ell} &= \text{LN}\Big(\mathbf{X}_{\ell-1} + \text{MHA}(\mathbf{X}_{\ell-1})\Big), \\
    \mathbf{X}''_{\ell} &= \text{LN}\Big(\mathbf{X}'_{\ell} + \text{CrossMHA}(\mathbf{X}'_{\ell},\, \mathbf{C})\Big), \\
    \mathbf{X}_{\ell} &= \text{LN}\Big(\mathbf{X}''_{\ell} + \text{FFN}(\mathbf{X}''_{\ell})\Big),
\end{align}
where $\text{MHA}$ is multi-head self-attention over the ligand sequence (with RoPE), $\text{CrossMHA}$ is multi-head cross-attention into $\mathbf{C}$, and $\text{FFN}(x)=W_2\,\phi(W_1x+b_1)+b_2$ with nonlinearity $\phi$.

The final hidden states are projected to vocabulary logits to predict token distributions required by the masked diffusion objective.

\subsection{Training Objective}
We optimize a masked diffusion objective on SMILES tokens with noise level $\sigma(t)$:
\begin{align}
\mathcal{L}_{\text{MDLM}} &= -\mathbb{E}_{x_0,t}\,\log p_\theta(x_0 \mid x_t, \mathcal{P}) \cdot \frac{d\sigma/dt}{\exp(\sigma)-1}, \\
x_t &\sim q(x_t \mid x_0, \sigma(t)),
\end{align}
where $x_t$ is produced by masking tokens with probability $1-\exp(-\sigma(t))$ and $\mathcal{P}$ is the conditioning context.

\subsection{Forward Process (Discrete Masking)}
Let $m(t)=1-\exp(-\sigma(t))$ be the mask probability. For each token $x_0^{(i)}$, we sample
\begin{align}
q(x_t^{(i)}=\texttt{[MASK]}\mid x_0^{(i)}) &= m(t), \\
q(x_t^{(i)}=x_0^{(i)}\mid x_0^{(i)}) &= 1-m(t),
\end{align}
and apply this independently over positions. The resulting $x_t$ is embedded and passed to the decoder.

\subsection{SUBS Parameterization}
We use the substitution (SUBS) parameterization: logits for the mask token are set to $-\infty$ and unmasked positions are constrained to remain unchanged,
\begin{align}
\log p_\theta(x_{t-1}\mid x_t) &\leftarrow \log p_\theta(x_{t-1}\mid x_t) - \log Z, \\ 
\log p_\theta(x_{t-1}=x_t\mid x_t\neq \texttt{[MASK]}) &= 0.
\end{align}

\subsection{Training Algorithm (High Level)}
Given batch $(\mathcal{P}, x_0)$, we sample $t\sim\mathcal{U}[0,1]$, mask $x_0$ to obtain $x_t$, run the decoder with conditioning $\mathcal{P}$, and minimize $\mathcal{L}_{\text{MDLM}}$ with AdamW. Mixed precision (bf16) and gradient accumulation are used for efficiency.

\begin{algorithm}[h]
\caption{Training (Masked Diffusion)}
\begin{algorithmic}[1]
\STATE Sample $t\sim \mathcal{U}(0,1)$ and compute $\sigma(t)$
\STATE Mask tokens with probability $m(t)=1-\exp(-\sigma(t))$ to obtain $x_t$
\STATE Forward: $(\hat{x}_0,\cdot)=f_\theta(\mathcal{P}, x_t, \sigma(t))$
\STATE Compute $\mathcal{L}_{\text{MDLM}}$ and backpropagate with AdamW (bf16 + grad accumulation)
\end{algorithmic}
\end{algorithm}

\subsection{Sampling Algorithm}
At inference, we iterate $t$ from 1 to $\epsilon$ in $T$ steps, denoise $x_t$ with the decoder, and sample the next tokens using nucleus sampling ($p=0.9$). A final noise-removal pass produces $x_0$.

\begin{algorithm}[h]
\caption{Sampling (ReMDM-cap + SUBS)}
\begin{algorithmic}[1]
\STATE Initialize $x_T \leftarrow \texttt{[MASK]}$ tokens
\STATE Set a timestep grid $t_0=1 > t_1 > \cdots > t_T=\epsilon$ with step size $\Delta t$
\FOR{$i=1$ to $T$}
    \STATE Compute $\sigma_i = \sigma(t_i)$ and mask probability $m_i = 1-\exp(-\sigma_i)$
    \STATE Predict normalized log-probabilities $\log p_\theta(x_0\mid x_{t_i})$ and apply SUBS parameterization (mask token suppressed; unmasked tokens fixed)
    \STATE Convert to $p_\theta(x_0\mid x_{t_i})$ (apply temperature schedule and nucleus top-$p$)
    \STATE Let $\alpha_i = 1-m_i$ and $\alpha_{i-1}=1-m_{i-1}$ and set the capped noise $\tilde{\eta}=\min\!\left(\eta,\;\frac{1-\alpha_{i-1}}{\alpha_i}\right)$
    \STATE Form the categorical proposal $q(x_{t_{i-1}}\mid x_{t_i})$ by mixing $p_\theta(x_0\mid x_{t_i})$ with the mask token at rate $\tilde{\eta}$ (ReMDM-cap)
    \STATE Sample $x_{t_{i-1}} \sim q(\cdot\mid x_{t_i})$
\ENDFOR
\STATE Noise-removal step at $t=\epsilon$
\end{algorithmic}
\end{algorithm}

\section{Triangle Operators}
\label{app:triangle_ops}

We provide detailed formulations of the triangle attention and multiplication operators applied to coarse pair features $\mathbf{p}_c \in \mathbb{R}^{B \times K \times K \times C}$. Each operator follows the same pattern: (i) layer-normalize the pair features, (ii) project into head space, (iii) apply attention or bilinear interaction, (iv) gate and project back to $C$.

\subsection{Triangle Attention}

\textbf{Starting Node Update:}
\begin{align}
\tilde{\mathbf{p}} &= \text{LN}(\mathbf{p}_c), \\
\mathbf{q}_{i} &= W_Q\,\tilde{\mathbf{p}}_{i,:}, \quad \mathbf{k}_{i} = W_K\,\tilde{\mathbf{p}}_{i,:}, \quad \mathbf{v}_{i} = W_V\,\tilde{\mathbf{p}}_{i,:}, \\
\alpha_{i,j,k} &= \text{softmax}_k\!\left(\frac{\langle \mathbf{q}_{i,j}, \mathbf{k}_{i,k} \rangle}{\sqrt{d_h}} + b_{i,j,k}\right), \\
\mathbf{o}_{i,j} &= \sum_{k} \alpha_{i,j,k}\, \mathbf{v}_{i,k}, \\
\mathbf{p}_c[b,i,j] &\leftarrow W_O\big(\sigma(W_G\tilde{\mathbf{p}}_{i,j}) \odot \mathbf{o}_{i,j}\big)
\end{align}
For fixed batch $b$ and row $i$, each edge $(i,j)$ attends to all edges $(i,k)$ sharing the starting node $i$. The bias $b_{i,j,k}$ is derived from pair features and a mask.

\textbf{Ending Node Update:}
\begin{align}
\mathbf{p}_c[b,i,j] \leftarrow \text{TriAttn}_{\text{end}}\big(\mathbf{p}_c[b,\cdot, j]\big)
\end{align}
For fixed batch $b$ and column $j$, each edge $(i,j)$ attends to all edges $(k,j)$ sharing the ending node $j$.

\subsection{Triangle Multiplication}

\textbf{Outgoing Edges:}
\begin{align}
\tilde{\mathbf{p}} &= \text{LN}(\mathbf{p}_c), \\
\mathbf{a}_{i,k} &= W_A\,\tilde{\mathbf{p}}_{i,k}, \quad \mathbf{b}_{j,k} = W_B\,\tilde{\mathbf{p}}_{j,k}, \\
\mathbf{p}_c[b,i,j] &\leftarrow \sigma(W_G\tilde{\mathbf{p}}_{i,j}) \odot W_O\Big(\sum_{k=1}^K \mathbf{a}_{i,k} \odot \mathbf{b}_{j,k}\Big)
\end{align}
where $g$ and $h$ are linear projections. This aggregates information over paths $i \to k \gets j$.

\textbf{Incoming Edges:}
\begin{align}
\mathbf{p}_c[b,i,j] \leftarrow \sigma(W_G\tilde{\mathbf{p}}_{i,j}) \odot W_O\Big(\sum_{k=1}^K \mathbf{a}_{k,i} \odot \mathbf{b}_{k,j}\Big)
\end{align}
This aggregates information over paths $i \gets k \to j$.

These operations enforce geometric consistency in the learned pairwise features by leveraging triangle inequality constraints, similar to AlphaFold's protein structure prediction.

\subsection{Transition Factor Usage}
The \texttt{transition\_factor} controls the width expansion in the single and pair transition MLPs :
\begin{align}
\text{SingleFC}:\; \mathbb{R}^{C_{\text{single}}} \to \mathbb{R}^{r\,C_{\text{single}}} \to \mathbb{R}^{C_{\text{single}}}, \\ 
\text{PairFC}:\; \mathbb{R}^{C_{\text{pair}}} \to \mathbb{R}^{r\,C_{\text{pair}}} \to \mathbb{R}^{C_{\text{pair}}},
\end{align}

\subsection{Operator Masking and Stability}
We apply a pairwise mask $M \in \{0,1\}^{K \times K}$ derived from sequence padding. For attention, invalid keys are masked by adding $-\infty$ to logits. For multiplication, invalid entries are zeroed before aggregation. Gating terms $\sigma(W_G\cdot)$ stabilize training by limiting updates to confident edges.

\subsection{FoldingBlock Algorithm}
We follow the implementation order in {FoldingBlock}:
\begin{algorithm}[h]
\caption{FoldingBlock (coarse grid)}
\begin{algorithmic}[1]
\STATE $M \leftarrow$ mask; $\mathbf{s},\mathbf{p}$ are single/pair features
\STATE $\mathbf{s} \leftarrow \mathbf{s} + \text{SingleAttn}(\mathbf{s}, M, \text{attn\_bias}(\mathbf{p}))$
\STATE $\mathbf{s} \leftarrow \mathbf{s} + \text{SingleFC}(\mathbf{s};\; r=\texttt{transition\_factor})$
\STATE $\mathbf{p} \leftarrow \mathbf{p} + \text{OuterLinear}(\mathbf{s})$
\STATE $\mathbf{p} \leftarrow \mathbf{p} + \text{TriMult}_{\text{out}}(\mathbf{p}, M)$
\STATE $\mathbf{p} \leftarrow \mathbf{p} + \text{TriMult}_{\text{in}}(\mathbf{p}, M)$
\STATE $\mathbf{p} \leftarrow \mathbf{p} + \text{TriAttn}_{\text{start}}(\mathbf{p}, M)$
\STATE $\mathbf{p} \leftarrow \mathbf{p} + \text{TriAttn}_{\text{end}}(\mathbf{p}, M)$
\STATE $\mathbf{p} \leftarrow \mathbf{p} + \text{PairFC}(\mathbf{p};\; r=\texttt{transition\_factor})$
\STATE \textbf{return} $\mathbf{s},\mathbf{p}$
\end{algorithmic}
\end{algorithm}

\end{document}